%% file: acl_latex.tex
\pdfoutput=1
\documentclass[11pt]{article}

\usepackage[final]{acl}

\usepackage{times}
\usepackage{latexsym}

\usepackage[T1]{fontenc}
\usepackage[utf8]{inputenc}

\usepackage{microtype}
\usepackage{float} 
\usepackage{inconsolata}
\usepackage{multirow}
\usepackage{graphicx}
\usepackage{tikz}
\usetikzlibrary{shapes.geometric, arrows}

\tikzstyle{startstop} = [rectangle, rounded corners, minimum width=3cm, minimum height=1cm, text centered, draw=black, fill=white!30, font=\large]
\tikzstyle{process} = [rectangle, minimum width=3cm, minimum height=1cm, text centered, draw=black, fill=white!30, font=\large]
\tikzstyle{decision} = [diamond, minimum width=3cm, minimum height=1cm, text centered, draw=black, fill=white!30, font=\large]
\tikzstyle{arrow} = [thick,->,>=stealth]

\newcommand\anonymize[1]{[ANONYMIZED]}

\usepackage[normalem]{ulem}

\usepackage{tipa}
\usepackage{xcolor,colortbl}
\usepackage{float}
\usepackage{inconsolata}
\usepackage{pgfplots}
\usepackage{geometry}
\definecolor{Gray}{gray}{0.85}

\definecolor{peru22113282}{RGB}{221,132,82}
\definecolor{steelblue76114176}{RGB}{76,114,176}

\definecolor{crimson2152528}{RGB}{215,25,28}
\definecolor{darkgray176}{RGB}{176,176,176}
\definecolor{lightgray204}{RGB}{204,204,204}
\definecolor{steelblue44123182}{RGB}{44,123,182}

\pgfplotsset{compat=1.18} 

\title{Frequency matters: Modeling irregular morphological patterns in Spanish with Transformers}

\author{
\textbf{Akhilesh Kakolu Ramarao}$^1$, \textbf{Kevin Tang}$^{1,3}$, \textbf{Dinah Baer-Henney}$^2$ \\
Faculty of Arts and Humanities, Heinrich Heine University Düsseldorf \\
$^1$Department of English Language and Linguistics \hspace{1em} $^2$Institute of Linguistics,\\
$^3$Department of Linguistics, College of Liberal Arts and Sciences, University of Florida\\
\texttt{\{akhilesh.kakolu.ramarao, kevin.tang, dinah.baer-henney\}@uni-duesseldorf.de}
}

\begin{document}
\maketitle
\begin{abstract}

Over the past decade, various studies have addressed how speakers solve the so-called `The Paradigm Cell Filling Problem' (PCFP) \citep{ackerman2009parts} across different languages. The PCFP addresses a fundamental question in morphological processing: how do speakers accurately generate inflected forms of words when presented with incomplete paradigms? This problem is particularly salient when modeling complex inflectional systems. We focus on Spanish verbal paradigms, where certain verbs follow an irregular L-shaped pattern, where the first-person singular present indicative stem matches the stem used throughout the present subjunctive mood. We formulate the problem as a morphological reinflection task. Specifically, we investigate the role of input frequency in the acquisition of regular versus irregular L-shaped patterns in transformer models. By systematically manipulating the input distributions and analyzing model behavior, we reveal four key findings: 1) Models perform better on L-shaped verbs compared to regular verbs, especially in uneven frequency conditions; 2) Robust primacy effects are observed, but no consistent recency effects; 3) Memorization becomes more prominent as the proportion of L-shaped verbs increases; 4) There is a tendency to regularize L-shaped verbs when their consonant alternation pairs are rare or absent in the training data.

\end{abstract}

\section{Introduction}

A common generation task in morphology is morphological inflection, where a target form has to be generated from its corresponding lemma and feature tag, e.g., \texttt{(lemma:decir, target tag:<V;IND;PRS;1;SG>)} $\mapsto$ \texttt{digo}. A central challenge in understanding how speakers handle morphological inflection is the Paradigm Cell Filling Problem (PCFP) \citep{ackerman2009parts}, which asks how speakers can reliably produce inflected forms of words when they are presented with incomplete paradigms.

To address the PCFP, encoder-decoder based neural networks have been used to simulate the learning and generation of inflected forms \citep{silfverberg-hulden-2018-encoder, wiemerslage2022comprehensive}. Our study extends this line of research by applying the PCFP to a morphomic pattern using encoder-decoder transformers.  
The morphomic pattern, as introduced by \citet{Aronoff-1994}, is a morphological pattern that exists independently of semantics or syntax. It is purely based on the form and structure of words. A key characteristic of morphomic patterns is their predictability within the verbal paradigm. The verb forms that are part of the pattern share morphological features, despite a lack of apparent semantic or syntactic motivation \citep{blevins2016word,maiden2018romance}. 

\citet{maiden2011allomorphy,maiden2018romance,maiden2021morphome} identified morphomic patterns across Romance languages. We focus on Spanish for data availability reasons \citep{herce2024meaning}. To illustrate an example, the Spanish verb forms ``digo'' (1st person singular, indicative) and ``digan'' (3rd person plural, subjunctive) of the verb \textit{decir} `to say'  (see Table \ref{tab: decir-example}) share the stem ``dig-''. This shared morphological feature is part of a morphomic pattern. However, there is no obvious semantic or syntactic property that links ``digo'' and ``digan'' while excluding ``dicen'' (3rd person plural, indicative) of the same verb, which uses a different stem ``dic-''. 

Spanish exhibits several morphomic patterns, namely L-, N-, P- and F-shaped patterns \citep{maiden2018romance,herce2024meaning}. Of these, only the L-shaped pattern has been explored for human learnability \citep{Nevins2015TheRA, cappellaro2024cognitive}. We choose the L-shaped pattern for our study because it allows us to assess the cognitive plausibility of our neural network models. The L-shaped pattern is characterized by the use of a distinct stem form in the first person singular present indicative and all cells of the present subjunctive mood. For example, the irregular verb \textit{decir} exhibits the L-shaped morphome pattern, as shown in Table \ref{tab: decir-example}.

\begin{table}[H]
\centering
\resizebox{\columnwidth}{!}{%
\begin{tabular}{lllll}
\hline
\textbf{`to say'} & \multicolumn{2}{c}{\textbf{\textit{Indicative}}} & \multicolumn{2}{c}{\textbf{\textit{Subjunctive}}} \\
\hline
& Orthographic & IPA & Orthographic & IPA \\
\hline
\textbf{\textit{1SG}} & \cellcolor[HTML]{FFCE93}digo & \cellcolor[HTML]{FFCE93}d\textipa{"}i\textipa{g}o & \cellcolor[HTML]{FFCE93} diga & \cellcolor[HTML]{FFCE93} d\textipa{"}i\textipa{g}a \\
\textbf{\textit{2SG}} & dices & d\textipa{"}ises & \cellcolor[HTML]{FFCE93} digas & \cellcolor[HTML]{FFCE93} d\textipa{"}i\textipa{g}as \\
\textbf{\textit{3SG}} & dice & d\textipa{"}ise & \cellcolor[HTML]{FFCE93} diga & \cellcolor[HTML]{FFCE93} d\textipa{"}i\textipa{g}a \\
\textbf{\textit{1PL}} & decimos & des\textipa{"}imos & \cellcolor[HTML]{FFCE93} digamos & \cellcolor[HTML]{FFCE93} di\textipa{g"}amos \\
\textbf{\textit{2PL}} & decís & des\textipa{"}is & \cellcolor[HTML]{FFCE93} digáis & \cellcolor[HTML]{FFCE93} di\textipa{g"}ajs \\
\textbf{\textit{3PL}} & dicen & d\textipa{"}isen & \cellcolor[HTML]{FFCE93} digan & \cellcolor[HTML]{FFCE93} d\textipa{"}i\textipa{g}an \\
\hline
\end{tabular}
}
\caption{A Spanish example of the Romance L-pattern, verb \textit{decir} 'to say'. L-shaped pattern cells are shaded.
}
\label{tab: decir-example}
\end{table}

Spanish L-shaped verbs exhibit an interesting distribution in the lexicon: they are found in relatively few word types but demonstrate high token frequency, which is the frequency of occurrence of individual word forms \citep{maiden2011allomorphy}. The role of type frequency, which refers to the number of different words that follow a particular morphological pattern in morphological productivity has been well established in the linguistic literature  \citep{bybee1995,pierrehumbert2001stochastic,bybee2003phonology,Albright2003,baer2012role,del2004putting}. These studies show that patterns with higher type frequency are more likely to be extended to novel forms, suggesting a strong correlation between type frequency and productivity. However, previous studies on L-shaped verbs challenge this established relationship. \citet{Nevins2015TheRA} and \citet{cappellaro2024cognitive} find conflicting evidence on productivity of morphomes in human studies. This highlights the need for a computational approach that can systematically explore the factors influencing morphome productivity. Computational modeling allows us to manipulate type frequency in ways that would be challenging in human studies, enabling a more controlled investigation of its effects on morphomic pattern learnability and productivity. 

In our study, we investigate the impact of type frequency on the learnability of morphomic patterns, specifically the L-shaped pattern, by implementing a morphological reinflection task framed as a PCFP using transformer models. 

The morphological reinflection task aligns with the morphological framework of abstraction based on data directly available to speakers (i.e., inflection forms) \citep{blevins2006,Boye2019}, providing a realistic setting.
We implement a multi-source setup of the morphological reinflection task \citep{Kann_etal_2017_EACL}, which uses multiple source form-tag pairs instead of one form-tag pair. The task here is to generate an inflected form from two source form-tag pairs and the target feature tag to predict the target inflected form, e.g., \texttt{(source form 1:digo, source tag 1:<V;IND;PRS;1;SG>, source form 2:diga, source tag 2:<V;SBJV;PRS;1;SG>, target tag:<V;SBJV;PRS;2;SG>)} $\mapsto$ \texttt{digas}. The choice of two-source setup is motivated by two key considerations: 1) identifying L-shaped verbs requires knowledge of at least two paradigm cells (one cell within the L-shaped pattern, one cell outside; see Table \ref{tab: decir-example}), and 2) previous research shows that two-source form-tag pairs are sufficient for achieving high accuracy in paradigm completion, with no gains from additional sources \citep{SilfverbergH18,LiuH20}. 

The reinflection task is particularly challenging due to the variability of the starting point (the source), which can be any other inflected form of the same lemma. This variability makes the task more cognitively plausible, reflecting the data sparsity encountered by human speakers, who never encounter all of the inflected forms of their language \citep{Blevins2017}.

The main aims of the study are: 1) To model the learning of the L-shaped morphome in Spanish using transformers for morphological reinflection; 2) To analyze the models' performance across varying input frequency distributions of regular and L-shaped verbs; 3) To conduct post-hoc analyses investigating: a) the impact of paradigm cell combinations on L-shaped learning, b) models' ability to memorize and generalize morphomic patterns, and c) sensitivity to the input frequency of consonant alternations.

We publish the dataset and code used in our study at \url{https://github.com/hhuslamlab/modeling_spanish_acl}

\section{Related Work}\label{sec:related-work}

\citet{silfverberg-hulden-2018-encoder} introduced the encoder-decoder approach to PCFP by formulating the problem as a morphological reinflection task. The following paragraphs will provide an overview of the methodologies used to address this problem over time. 

Initial non-neural models focused on learning string edit rules from data using sequence-to-sequence models \citep{Albright2003,durrett2013supervised} and string transductions \citep{nicolai2015inflection}. Subsequently, \citet{ahlberg2015paradigm} proposed a finite state construction and used a classifier to select the correct inflection. Similarly, \citet{alegria-etxeberria-2016-ehu} used a single Weighted Finite-State Transducer (WFST) \citep{novak-etal-2012-wfst} to model the mapping between lemmas and inflected forms.  \citet{taji2016columbia} developed a morphological analyzer that learns from the training data and applies the learned patterns to re-inflect test data. \citet{nicolai-etal-2016-morphological} implemented a discriminative transducer based on \citet{jiampojamarn-etal-2008-joint} that searches for a series of character transformation rules to perform the inflection. \citet{liu-mao-2016-morphological} and \citet{cotterell-etal-2017-conll} applied affixing rules to generate inflected forms, which involved appending or altering affixes (prefixes, suffixes, infixes, etc.) according to morphological rules. More recently, \citet{sherbakov2022morphology} used a non-neural system to predict inflected forms based on string patterns observed in training samples. \citet{kwak2023morphological} introduced an improved affixing system that incorporated additional linguistic information to better capture the complexities in morphological generation. These methods, while interpretable, often struggled with irregular forms and low-resource scenarios.

The introduction of neural-based sequence-to-sequence models marked a significant milestone in modeling morphological reinflection by learning complex morphological patterns without explicit rules \citep{kann2016single,malouf2016generating,faruqui2015morphological}. Subsequent studies built upon this approach, including the hard monotonic attention for strict alignment between input and output sequences \citep{wu2019exact}, multi-source setups with bi-directional LSTMs \citep{kann2016single}, character-level LSTMs \citep{silfverberg-hulden-2018-encoder}, and encoder-decoder based transformers \citep{Wu2021ApplyingTT}. Recent developments include phonologically-aware embeddings \citep{guriel2023morphological} to capture both orthographic and phonetic information of words. Other approaches include treating morphological reinflection as a classification problem \citep{shcherbakov2023does}, and the application of imitation learning \citep{makarov2018imitation}. Lastly, Large Language Models have also been explored for morphological reinflection, including analyzing ChatGPT's capability in morphological generation across multiple languages \citep{weissweiler2023counting}. 

Our study implements a model that closely follows the formulation of the encoder-decoder transformer for character-level transduction proposed by \citet{Wu2021ApplyingTT}, due to its high performance on inflectional tasks across various languages \citep{cotterell-etal-2017-conll, cotterell-etal-2018-conll, vylomova-etal-2020-sigmorphon}. 

\section{Methodology}\label{sec:methodology}

\subsection{Model Architecture} 

We implement the model using \texttt{fairseq} \cite{ott2019fairseq}, a PyTorch-based sequence modeling toolkit. The model consists of four layers with four attention heads, an embedding size of 256, and a hidden layer size of 1,024. We use the Adam Optimizer \cite{Kingma2015AdamAM} with an initial learning rate of 0.001, 0.1 label smoothing, and a 1.0 gradient clip threshold. The model is trained for a maximum of 10,000 optimizer updates, with checkpoints saved every ten epochs. Beam search is used at decoding time with a beam width of five.

\textbf{Hyperparameter tuning} Hyperparameter tuning, particularly the batch size, plays a crucial role in seq2seq tasks \citep{Wu2021ApplyingTT,popel2018training}. We use varying batch sizes, from 32 to 3,600, and observed that the impact of batch size on the accuracy of predicting L-shaped verbs is not uniform across frequency conditions (see Figure \ref{fig:batch-sizes} in Appendix). We adopt a batch size of 400 following established practices in morphological tasks \citep{vylomova-etal-2020-sigmorphon,pimentel-ryskina-etal-2021-sigmorphon,kodner-khalifa-2022-sigmorphon}.

\subsection{Dataset construction} 

We use the Spanish verbal morphology dataset from the Universal Morphology (UniMorph) project\footnote{https://unimorph.github.io/}. The entries in the dataset are coded in the Unimorph scheme \citep{Sylak-Glassman2016}. For example, the label   
\texttt{V;IND;PRS;1;SG} , corresponding to a first person singular present tense verb form, such as \texttt{digo}, is decomposed into a set of morphosyntactic features: [\/POS=\textsc{verb}, mood=\textsc{indicative},
tense=\textsc{present}, person=\textsc{1}, number=\textsc{singular}]. The data representation in UniMorph follows the structure: (\texttt{lemma, form, feature}). We transcribe the lemma and inflected form in the International Phonetic Alphabet (IPA) to capture phonological representations, resulting in an entry such as (\texttt{\textipa{desiR}, \textipa{digo}, V;IND;PRS;1;SG}). 

For the reinflection task, we convert these entries to two source form-tag pairs: the target feature tag and the target inflected form. For example, if the first source form is \texttt{\textipa{digo}} (1st person singular, indicative), the second source form is \texttt{\textipa{diga}} (1st person singular, subjunctive) and the target form is \texttt{\textipa{digas}} (2nd person singular, subjunctive), then the above entry is converted to a so-called \textit{triple} entry such as (\texttt{\textipa{digo}, V;IND;PRS;1;SG, \textipa{diga}, V;SBJV;PRS;1;SG, V;SBJV;PRS;2;SG, \textipa{digas}}). The dataset contains 5,460 distinct lemmas, of which 300 are L-shaped lemmas, and 4,860 are NL-shaped lemmas, which results in 382,956 triples. 

To investigate the role of input frequency, we implement three experimental conditions, each characterized by a different ratio of L-shaped to regular (henceforth, \textit{NL-shaped}) verbs in the training set under a) a naturalistic frequency distribution with 10\% L-shaped verbs and 90\% NL-shaped verbs (henceforth, \textit{10\%L-90\%NL condition}) to reflect a realistic frequency distribution of the Spanish language\footnote{This is similar to the relative frequencies of L and NL-shaped verbs in the dataset which is 6\% L and 94\% NL.}, and two counterfactual conditions with an increase in the frequency of L-shaped verbs, and a decrease in the frequency of the NL-shaped verbs: 
b) 50\% L-shaped verbs and 50\% NL-shaped verbs (henceforth, \textit{50\%L-50\%NL condition}) and c) 90\% L-shaped verbs and 10\% NL-shaped verbs (henceforth, \textit{90\%L-10\%NL condition}). The relative frequency of these counterfactual conditions is created to allow a direct comparison of the learnability of L-shaped verbs relative to NL-shaped verbs. 

\textbf{Data representation} The input sequence for our model is structured as follows: 

\texttt{
\textipa{d i g a} \# <V;SBJV;PRS;1;SG> \# \textipa{d i g a s} \# 
<V;SBJV;PRS;2;SG> \# <V;IND;PRS;1;SG>}

The expected target output is the space-separated characters forming the target word, \texttt{\textipa{d i g o}}. We refer to this input-output sequence as a \textit{combination}. 

\textbf{Data sampling} To isolate the effect of relative type frequency, we used an identical set of lemmas across all three conditions. As mentioned above, only 300 L-shaped lemmas are found in the UniMorph dataset, therefore, the maximum number of L-shaped lemmas in the 90\%L-10\%NL condition is capped to 300, representing 90\%L. Therefore, the training set contains 333 lemmas, and we need to sample 33 NL-shaped verbs (amongst the  4,860 NL-shaped verbs) to represent 10\% of NL. Similarly, we sample for the other two conditions (50\%L-50\%NL and 10\%L-90\%NL).

We implement a rigorous data splitting strategy to mitigate the risk of artificially inflated model performance due to lemma overlap between training and testing data \citep{Goldman2022UnsolvingMI, kodner2023morphological}. At the \textit{lemma level}, we ensure no lemma overlap between the training, development, and test sets. We apply a 70-10-20 split ratio to the lemmas, with 70\% (training), 10\% (development), and 20\% (testing). At the \textit{combination level}, given that each lemma produces approximately 600 combinations and there are 333 lemmas in each condition (resulting in 199,800 combinations), we partition the data into four bins to manage computational complexity and maintain cognitive plausibility. Each bin preserves the condition-specific distribution of L and NL-shaped lemmas (e.g., 10\%L-90\%NL). Finally, at the \textit{run level}, we implement three randomized runs for each combination bin to account for potential order effects during training. Specifically, we generate three distinct training sets for each of the four combination bins created at the combination level.

The training data comprises full inflection tables, with which the model inflects unseen lemmas. For the development and test data, every two-slot combination of given slots is used as input to predict the target form corresponding to the target Morphosyntactic description (MSD) tag. Across all frequency conditions, our sampling procedure yields a training set of 39,435 combinations, a development set of 4,455 combinations, and a test set of 44,220 combinations. To enhance the robustness of our evaluation, we maintain a constant test set for each combination bin. In total, we generate 12 such datasets for each frequency condition. See Figure \ref{fig: condition-sampling} for an illustration of our data sampling procedure.

\begin{figure}[H]
\centering
\resizebox{\columnwidth}{!}{

\tikzset{every picture/.style={line width=0.75pt}} %set default line width to 0.75pt        

\begin{tikzpicture}[x=0.75pt,y=0.75pt,yscale=-1,xscale=1]
%uncomment if require: \path (0,745); %set diagram left start at 0, and has height of 745

%Straight Lines [id:da5945968684402243] 
\draw    (137.6,171) -- (138.98,193.5) ;
\draw [shift={(139.1,195.5)}, rotate = 266.5] [color={rgb, 255:red, 0; green, 0; blue, 0 }  ][line width=0.75]    (10.93,-3.29) .. controls (6.95,-1.4) and (3.31,-0.3) .. (0,0) .. controls (3.31,0.3) and (6.95,1.4) .. (10.93,3.29)   ;
%Straight Lines [id:da8273618662099527] 
\draw    (113.6,247.5) -- (204.86,247) ;
%Straight Lines [id:da6886481564280014] 
\draw    (159.6,247.5) -- (159.86,278) ;
\draw [shift={(159.88,280)}, rotate = 269.52] [color={rgb, 255:red, 0; green, 0; blue, 0 }  ][line width=0.75]    (10.93,-3.29) .. controls (6.95,-1.4) and (3.31,-0.3) .. (0,0) .. controls (3.31,0.3) and (6.95,1.4) .. (10.93,3.29)   ;
%Straight Lines [id:da9103179623133044] 
\draw    (204.86,247) -- (205.8,278) ;
\draw [shift={(205.86,280)}, rotate = 268.26] [color={rgb, 255:red, 0; green, 0; blue, 0 }  ][line width=0.75]    (10.93,-3.29) .. controls (6.95,-1.4) and (3.31,-0.3) .. (0,0) .. controls (3.31,0.3) and (6.95,1.4) .. (10.93,3.29)   ;
%Straight Lines [id:da8506317052391581] 
\draw    (292.6,171) -- (293.89,194) ;
\draw [shift={(294,196)}, rotate = 266.79] [color={rgb, 255:red, 0; green, 0; blue, 0 }  ][line width=0.75]    (10.93,-3.29) .. controls (6.95,-1.4) and (3.31,-0.3) .. (0,0) .. controls (3.31,0.3) and (6.95,1.4) .. (10.93,3.29)   ;
%Straight Lines [id:da47599303872216425] 
\draw    (444.6,171) -- (445.28,182.96) -- (445.77,195) ;
\draw [shift={(445.85,197)}, rotate = 267.66] [color={rgb, 255:red, 0; green, 0; blue, 0 }  ][line width=0.75]    (10.93,-3.29) .. controls (6.95,-1.4) and (3.31,-0.3) .. (0,0) .. controls (3.31,0.3) and (6.95,1.4) .. (10.93,3.29)   ;
%Straight Lines [id:da954073825920541] 
\draw    (591.6,171) -- (592.28,182.96) -- (592.99,195.5) ;
\draw [shift={(593.1,197.5)}, rotate = 266.76] [color={rgb, 255:red, 0; green, 0; blue, 0 }  ][line width=0.75]    (10.93,-3.29) .. controls (6.95,-1.4) and (3.31,-0.3) .. (0,0) .. controls (3.31,0.3) and (6.95,1.4) .. (10.93,3.29)   ;
%Straight Lines [id:da16076739355920577] 
\draw    (137.6,171) -- (591.6,171) ;
%Straight Lines [id:da5381651319685103] 
\draw    (160,228) -- (159.6,236) -- (159.6,247.5) ;
%Straight Lines [id:da7513271604315341] 
\draw    (113.6,247.5) -- (113.86,278) ;
\draw [shift={(113.88,280)}, rotate = 269.52] [color={rgb, 255:red, 0; green, 0; blue, 0 }  ][line width=0.75]    (10.93,-3.29) .. controls (6.95,-1.4) and (3.31,-0.3) .. (0,0) .. controls (3.31,0.3) and (6.95,1.4) .. (10.93,3.29)   ;
%Rounded Rect [id:dp48320523888575484] 
\draw   (280,16.4) .. controls (280,8.45) and (286.45,2) .. (294.4,2) -- (420.6,2) .. controls (428.55,2) and (435,8.45) .. (435,16.4) -- (435,59.6) .. controls (435,67.55) and (428.55,74) .. (420.6,74) -- (294.4,74) .. controls (286.45,74) and (280,67.55) .. (280,59.6) -- cycle ;
%Shape: Rectangle [id:dp21229968179465808] 
\draw   (280,99) -- (460,99) -- (460,139) -- (280,139) -- cycle ;
%Rounded Rect [id:dp9226296810389527] 
\draw   (98,203.09) .. controls (98,199.72) and (100.72,197) .. (104.09,197) -- (215.63,197) .. controls (218.99,197) and (221.71,199.72) .. (221.71,203.09) -- (221.71,221.34) .. controls (221.71,224.7) and (218.99,227.43) .. (215.63,227.43) -- (104.09,227.43) .. controls (100.72,227.43) and (98,224.7) .. (98,221.34) -- cycle ;
%Rounded Rect [id:dp67303552653594] 
\draw   (242,203.09) .. controls (242,199.72) and (244.72,197) .. (248.09,197) -- (359.63,197) .. controls (362.99,197) and (365.71,199.72) .. (365.71,203.09) -- (365.71,221.34) .. controls (365.71,224.7) and (362.99,227.43) .. (359.63,227.43) -- (248.09,227.43) .. controls (244.72,227.43) and (242,224.7) .. (242,221.34) -- cycle ;
%Rounded Rect [id:dp052529951123044816] 
\draw   (381,204.09) .. controls (381,200.72) and (383.72,198) .. (387.09,198) -- (498.63,198) .. controls (501.99,198) and (504.71,200.72) .. (504.71,204.09) -- (504.71,222.34) .. controls (504.71,225.7) and (501.99,228.43) .. (498.63,228.43) -- (387.09,228.43) .. controls (383.72,228.43) and (381,225.7) .. (381,222.34) -- cycle ;
%Rounded Rect [id:dp42108115458433226] 
\draw   (525,204.09) .. controls (525,200.72) and (527.72,198) .. (531.09,198) -- (642.63,198) .. controls (645.99,198) and (648.71,200.72) .. (648.71,204.09) -- (648.71,222.34) .. controls (648.71,225.7) and (645.99,228.43) .. (642.63,228.43) -- (531.09,228.43) .. controls (527.72,228.43) and (525,225.7) .. (525,222.34) -- cycle ;
%Straight Lines [id:da12422742455819291] 
\draw    (365.71,140.71) -- (365.71,170.71) ;
%Straight Lines [id:da4348749581645892] 
\draw    (364,74.86) -- (364.91,95.86) ;
\draw [shift={(365,97.86)}, rotate = 267.51] [color={rgb, 255:red, 0; green, 0; blue, 0 }  ][line width=0.75]    (10.93,-3.29) .. controls (6.95,-1.4) and (3.31,-0.3) .. (0,0) .. controls (3.31,0.3) and (6.95,1.4) .. (10.93,3.29)   ;
%Rounded Rect [id:dp4006522751452586] 
\draw   (92,288.09) .. controls (92,285.43) and (94.15,283.29) .. (96.8,283.29) -- (124.2,283.29) .. controls (126.85,283.29) and (129,285.43) .. (129,288.09) -- (129,302.49) .. controls (129,305.14) and (126.85,307.29) .. (124.2,307.29) -- (96.8,307.29) .. controls (94.15,307.29) and (92,305.14) .. (92,302.49) -- cycle ;
%Rounded Rect [id:dp7051500102123913] 
\draw   (140,288.09) .. controls (140,285.43) and (142.15,283.29) .. (144.8,283.29) -- (172.2,283.29) .. controls (174.85,283.29) and (177,285.43) .. (177,288.09) -- (177,302.49) .. controls (177,305.14) and (174.85,307.29) .. (172.2,307.29) -- (144.8,307.29) .. controls (142.15,307.29) and (140,305.14) .. (140,302.49) -- cycle ;
%Rounded Rect [id:dp9570970647836015] 
\draw   (187,288.09) .. controls (187,285.43) and (189.15,283.29) .. (191.8,283.29) -- (219.2,283.29) .. controls (221.85,283.29) and (224,285.43) .. (224,288.09) -- (224,302.49) .. controls (224,305.14) and (221.85,307.29) .. (219.2,307.29) -- (191.8,307.29) .. controls (189.15,307.29) and (187,305.14) .. (187,302.49) -- cycle ;
%Straight Lines [id:da4131842825811345] 
\draw    (257.6,247.5) -- (348.86,247) ;
%Straight Lines [id:da5329244730007994] 
\draw    (303.6,247.5) -- (303.86,278) ;
\draw [shift={(303.88,280)}, rotate = 269.52] [color={rgb, 255:red, 0; green, 0; blue, 0 }  ][line width=0.75]    (10.93,-3.29) .. controls (6.95,-1.4) and (3.31,-0.3) .. (0,0) .. controls (3.31,0.3) and (6.95,1.4) .. (10.93,3.29)   ;
%Straight Lines [id:da7253995351938249] 
\draw    (348.86,247) -- (349.8,278) ;
\draw [shift={(349.86,280)}, rotate = 268.26] [color={rgb, 255:red, 0; green, 0; blue, 0 }  ][line width=0.75]    (10.93,-3.29) .. controls (6.95,-1.4) and (3.31,-0.3) .. (0,0) .. controls (3.31,0.3) and (6.95,1.4) .. (10.93,3.29)   ;
%Straight Lines [id:da01820185289861964] 
\draw    (304,228) -- (303.6,236) -- (303.6,247.5) ;
%Straight Lines [id:da5518001543126743] 
\draw    (257.6,247.5) -- (257.86,278) ;
\draw [shift={(257.88,280)}, rotate = 269.52] [color={rgb, 255:red, 0; green, 0; blue, 0 }  ][line width=0.75]    (10.93,-3.29) .. controls (6.95,-1.4) and (3.31,-0.3) .. (0,0) .. controls (3.31,0.3) and (6.95,1.4) .. (10.93,3.29)   ;
%Rounded Rect [id:dp4368506550073552] 
\draw   (236,288.09) .. controls (236,285.43) and (238.15,283.29) .. (240.8,283.29) -- (268.2,283.29) .. controls (270.85,283.29) and (273,285.43) .. (273,288.09) -- (273,302.49) .. controls (273,305.14) and (270.85,307.29) .. (268.2,307.29) -- (240.8,307.29) .. controls (238.15,307.29) and (236,305.14) .. (236,302.49) -- cycle ;
%Rounded Rect [id:dp5314667721285258] 
\draw   (284,288.09) .. controls (284,285.43) and (286.15,283.29) .. (288.8,283.29) -- (316.2,283.29) .. controls (318.85,283.29) and (321,285.43) .. (321,288.09) -- (321,302.49) .. controls (321,305.14) and (318.85,307.29) .. (316.2,307.29) -- (288.8,307.29) .. controls (286.15,307.29) and (284,305.14) .. (284,302.49) -- cycle ;
%Rounded Rect [id:dp5295813532021758] 
\draw   (331,288.09) .. controls (331,285.43) and (333.15,283.29) .. (335.8,283.29) -- (363.2,283.29) .. controls (365.85,283.29) and (368,285.43) .. (368,288.09) -- (368,302.49) .. controls (368,305.14) and (365.85,307.29) .. (363.2,307.29) -- (335.8,307.29) .. controls (333.15,307.29) and (331,305.14) .. (331,302.49) -- cycle ;
%Straight Lines [id:da9433374811704731] 
\draw    (401.6,247.5) -- (492.86,247) ;
%Straight Lines [id:da21597791518720744] 
\draw    (447.6,247.5) -- (447.86,278) ;
\draw [shift={(447.88,280)}, rotate = 269.52] [color={rgb, 255:red, 0; green, 0; blue, 0 }  ][line width=0.75]    (10.93,-3.29) .. controls (6.95,-1.4) and (3.31,-0.3) .. (0,0) .. controls (3.31,0.3) and (6.95,1.4) .. (10.93,3.29)   ;
%Straight Lines [id:da027488083225436055] 
\draw    (492.86,247) -- (493.8,278) ;
\draw [shift={(493.86,280)}, rotate = 268.26] [color={rgb, 255:red, 0; green, 0; blue, 0 }  ][line width=0.75]    (10.93,-3.29) .. controls (6.95,-1.4) and (3.31,-0.3) .. (0,0) .. controls (3.31,0.3) and (6.95,1.4) .. (10.93,3.29)   ;
%Straight Lines [id:da4852594493328233] 
\draw    (448,228) -- (447.6,236) -- (447.6,247.5) ;
%Straight Lines [id:da4042927158571148] 
\draw    (401.6,247.5) -- (401.86,278) ;
\draw [shift={(401.88,280)}, rotate = 269.52] [color={rgb, 255:red, 0; green, 0; blue, 0 }  ][line width=0.75]    (10.93,-3.29) .. controls (6.95,-1.4) and (3.31,-0.3) .. (0,0) .. controls (3.31,0.3) and (6.95,1.4) .. (10.93,3.29)   ;
%Rounded Rect [id:dp7261082691338341] 
\draw   (380,288.09) .. controls (380,285.43) and (382.15,283.29) .. (384.8,283.29) -- (412.2,283.29) .. controls (414.85,283.29) and (417,285.43) .. (417,288.09) -- (417,302.49) .. controls (417,305.14) and (414.85,307.29) .. (412.2,307.29) -- (384.8,307.29) .. controls (382.15,307.29) and (380,305.14) .. (380,302.49) -- cycle ;
%Rounded Rect [id:dp6953859834925336] 
\draw   (428,288.09) .. controls (428,285.43) and (430.15,283.29) .. (432.8,283.29) -- (460.2,283.29) .. controls (462.85,283.29) and (465,285.43) .. (465,288.09) -- (465,302.49) .. controls (465,305.14) and (462.85,307.29) .. (460.2,307.29) -- (432.8,307.29) .. controls (430.15,307.29) and (428,305.14) .. (428,302.49) -- cycle ;
%Rounded Rect [id:dp10930825030173974] 
\draw   (475,288.09) .. controls (475,285.43) and (477.15,283.29) .. (479.8,283.29) -- (507.2,283.29) .. controls (509.85,283.29) and (512,285.43) .. (512,288.09) -- (512,302.49) .. controls (512,305.14) and (509.85,307.29) .. (507.2,307.29) -- (479.8,307.29) .. controls (477.15,307.29) and (475,305.14) .. (475,302.49) -- cycle ;
%Straight Lines [id:da9441796502302569] 
\draw    (544.6,247.5) -- (635.86,247) ;
%Straight Lines [id:da3789289259927253] 
\draw    (590.6,247.5) -- (590.86,278) ;
\draw [shift={(590.88,280)}, rotate = 269.52] [color={rgb, 255:red, 0; green, 0; blue, 0 }  ][line width=0.75]    (10.93,-3.29) .. controls (6.95,-1.4) and (3.31,-0.3) .. (0,0) .. controls (3.31,0.3) and (6.95,1.4) .. (10.93,3.29)   ;
%Straight Lines [id:da16309222274251067] 
\draw    (635.86,247) -- (636.8,278) ;
\draw [shift={(636.86,280)}, rotate = 268.26] [color={rgb, 255:red, 0; green, 0; blue, 0 }  ][line width=0.75]    (10.93,-3.29) .. controls (6.95,-1.4) and (3.31,-0.3) .. (0,0) .. controls (3.31,0.3) and (6.95,1.4) .. (10.93,3.29)   ;
%Straight Lines [id:da22808342549727123] 
\draw    (591,228) -- (590.6,236) -- (590.6,247.5) ;
%Straight Lines [id:da44787852514837234] 
\draw    (544.6,247.5) -- (544.86,278) ;
\draw [shift={(544.88,280)}, rotate = 269.52] [color={rgb, 255:red, 0; green, 0; blue, 0 }  ][line width=0.75]    (10.93,-3.29) .. controls (6.95,-1.4) and (3.31,-0.3) .. (0,0) .. controls (3.31,0.3) and (6.95,1.4) .. (10.93,3.29)   ;
%Rounded Rect [id:dp5520980086256047] 
\draw   (523,288.09) .. controls (523,285.43) and (525.15,283.29) .. (527.8,283.29) -- (555.2,283.29) .. controls (557.85,283.29) and (560,285.43) .. (560,288.09) -- (560,302.49) .. controls (560,305.14) and (557.85,307.29) .. (555.2,307.29) -- (527.8,307.29) .. controls (525.15,307.29) and (523,305.14) .. (523,302.49) -- cycle ;
%Rounded Rect [id:dp04989230038270476] 
\draw   (571,288.09) .. controls (571,285.43) and (573.15,283.29) .. (575.8,283.29) -- (603.2,283.29) .. controls (605.85,283.29) and (608,285.43) .. (608,288.09) -- (608,302.49) .. controls (608,305.14) and (605.85,307.29) .. (603.2,307.29) -- (575.8,307.29) .. controls (573.15,307.29) and (571,305.14) .. (571,302.49) -- cycle ;
%Rounded Rect [id:dp6718045670583375] 
\draw   (618,288.09) .. controls (618,285.43) and (620.15,283.29) .. (622.8,283.29) -- (650.2,283.29) .. controls (652.85,283.29) and (655,285.43) .. (655,288.09) -- (655,302.49) .. controls (655,305.14) and (652.85,307.29) .. (650.2,307.29) -- (622.8,307.29) .. controls (620.15,307.29) and (618,305.14) .. (618,302.49) -- cycle ;

% Text Node
\draw (100,206.09) node [anchor=north west][inner sep=0.75pt]   [align=left] {{\large Combi. bin 1} {\small (25\%)}};
% Text Node
\draw (244,206.09) node [anchor=north west][inner sep=0.75pt]   [align=left] {{\large Combi. bin 2} {\small (25\%)}};
% Text Node
\draw (383,207.09) node [anchor=north west][inner sep=0.75pt]   [align=left] {{\large Combi. bin 3} {\small (25\%)}};
% Text Node
\draw (527,207.09) node [anchor=north west][inner sep=0.75pt]   [align=left] {{\large Combi. bin 4} {\small (25\%)}};
% Text Node
\draw (139.8,286.29) node [anchor=north west][inner sep=0.75pt]   [align=left] {\large{R1.2}};
% Text Node
\draw (92.8,285.29) node [anchor=north west][inner sep=0.75pt]   [align=left] {\large{R1.1}};
% Text Node
\draw (284,15) node [anchor=north west][inner sep=0.75pt]   [align=left] {{\Large \textbf{ \ \ \ \ \ Condition}}\\{\large (e.g., 10\%L-90\%NL)}};
% Text Node
\draw (282,113) node [anchor=north west][inner sep=0.75pt]   [align=left] {\large{Combination level splitting}};
% Text Node
\draw (289,288) node [anchor=north west][inner sep=0.75pt]   [align=left] {};
% Text Node
\draw (336,287) node [anchor=north west][inner sep=0.75pt]   [align=left] {};
% Text Node
\draw (384,288) node [anchor=north west][inner sep=0.75pt]   [align=left] {};
% Text Node
\draw (187.8,285.29) node [anchor=north west][inner sep=0.75pt]   [align=left] {\large{R1.3}};
% Text Node
\draw (235.8,286.29) node [anchor=north west][inner sep=0.75pt]   [align=left] {\large{R2.1}};
% Text Node
\draw (283.8,286.29) node [anchor=north west][inner sep=0.75pt]   [align=left] {\large{R2.2}};
% Text Node
\draw (330.8,285.29) node [anchor=north west][inner sep=0.75pt]   [align=left] {\large{R2.3}};
% Text Node
\draw (381.8,285.29) node [anchor=north west][inner sep=0.75pt]   [align=left] {\large{R3.1}};
% Text Node
\draw (429.8,285.29) node [anchor=north west][inner sep=0.75pt]   [align=left] {\large{R3.2}};
% Text Node
\draw (475.8,285.29) node [anchor=north west][inner sep=0.75pt]   [align=left] {\large{R3.3}};
% Text Node
\draw (524.8,285.29) node [anchor=north west][inner sep=0.75pt]   [align=left] {\large{R4.1}};
% Text Node
\draw (571.8,285.29) node [anchor=north west][inner sep=0.75pt]   [align=left] {\large{R4.2}};
% Text Node
\draw (619,286.09) node [anchor=north west][inner sep=0.75pt]   [align=left] {\large{R4.3}};

\end{tikzpicture}

}
\caption{Flowchart showing the process for creating the dataset for each condition at the \textit{combination level}, and \textit{run level}.}
\label{fig: condition-sampling}
\end{figure}

% \textbf{Model Evaluation} We evaluate the accuracies of the predicted reinflection form across all three frequency conditions by comparing it with the ground truth. In particular, we consider the models' \textit{sequence accuracy}, where only instances for which the entire output which is in the form of a sequence of IPA symbols matches the target sequence are considered correct\footnote{Stress marks were ignored.}.

\section{Analysis and Results}

We begin by presenting the sequence accuracies of the models across the three frequency conditions. Subsequently, we conduct a series of post-hoc analyses to understand the factors influencing learning of morphomic patterns under varying frequency conditions. In Section \ref{sec:cellcombo}, we examine potential position biases in models. Section \ref{sec:memgen} investigates the models' sensitivity to memorization versus generalization effects \citep{hupkes2023taxonomy}. Lastly, in Section \ref{sec:consonant-pair-analysis}, we examine the models' behavior for specific phonological alternations.

We analyze the models' performance across varying input frequency distributions of regular and L-shaped verbs, addressing our research aim 2. We use two evaluation metrics: sequence accuracy and stem-only accuracy, as the errors can only occur in the suffixes and the stem of the predicted form \citep{kodner-khalifa-2022-sigmorphon}. 

We first evaluate sequence accuracies (refer to Figure \ref{fig:l-nl-overall-accuracies}) across frequency conditions. In the 10\%L-90\%NL condition, NL-shaped verbs have a mean accuracy of 61.8\%, which is 25.05\% higher than the 36.75\% mean accuracy of L-shaped verbs. Conversely, in the 90\%L-10\%NL condition, L-shaped verbs have a mean accuracy of 88.75\%, outperforming NL-shaped verbs by 64.58\%, which have a mean accuracy of 24.17\%. In the 50\%L-50\%NL condition, L-shaped verbs have a mean accuracy of 72.31\%, while NL-shaped verbs have a mean accuracy of 55.19\%, with L-shaped verbs performing 17.12\% better.

Subsequently, we analyze stem accuracies to isolate the models' performance on stem alternations across the frequency conditions (Appendix \ref{sec:appendix:accuracies}). The results show a clear trend: the more frequent verb type in each condition consistently has better stem accuracy.  
In the 10\%L-90\%NL condition, NL-shaped verbs achieve a mean stem accuracy of 68.59\%, which is 11.89\% higher than the 56.7\% mean stem accuracy of L-shaped verbs. In the 50\%L-50\%NL condition, L-shaped verbs have a mean stem accuracy of 77.24\%, slightly better than NL-shaped verbs at 70.25\%, with a difference of 6.99\%. In the 90\%L-10\%NL condition, L-shaped verbs have a higher mean stem accuracy of 89.51\%, outperforming NL-shaped verbs which have a mean stem accuracy of 31.09\%, showing a difference of 58.2\%.

% Next, we analysed the stem accuracies. The results show that L-shaped verbs had higher stem accuracies when they were more prevalent in the training data. In the 10\%L-90\%NL condition, NL-shaped verbs had a mean stem accuracy of 68.59\%, while L-shaped verbs had a mean stem accuracy of 56.7\%. 
% In the 50\%L-50\%NL condition, L-shaped verbs had a mean stem accuracy of 77.24\%, while NL-shaped verbs had a mean stem accuracy of 70.25\%. In the 90\%L-10\%NL condition, L-shaped verbs had a mean stem accuracy of 89.51\%, while NL-shaped verbs had a mean stem accuracy of 31.09\%, illustrated in Appendix \ref{sec:appendix:accuracies}.

These results show a clear performance difference between the models based on the distribution of verb types in the training data. In the 10\%L-90\%NL condition, which most closely approximates the natural distribution in Spanish, the models perform better on NL-shaped verbs than on L-shaped verbs. However, we find a learning advantage for L-shaped verbs in other conditions, suggesting the models might be learning specific characteristics of L-shaped verbs. To further understand the factors influencing the acquisition of morphomic patterns in the models', we conduct a series of post-hoc analyses.

\begin{figure}[h]
\centering
\scalebox{0.8}
  {\input{plots/lvsnl-accuracy-without-stress}}
    \caption{Mean and 95\% confidence intervals of overall sequence accuracies for L-shaped and NL-shaped verbs across frequency conditions (10\%L-90\%NL, 50\%L-50\%NL and 90\%L-10\%NL conditions).}
    \label{fig:l-nl-overall-accuracies}
\end{figure}

\subsection{Cell combinations}\label{sec:cellcombo}

We examine the influence of paradigm cell combinations on the acquisition of L-shaped verbs, in line with our research aim 3a. Our analysis reveal robust primacy effects but recency effects are inconsistent.

Transformers are prone to \textit{position bias}, disproportionately focusing on specific token positions due to  architectural constraints \citep{dufter2022}. We examine position bias in transformer through two psycholinguistically grounded metrics: primacy and recency effects. The primacy effect is a cognitive bias in which humans tend to remember and be influenced by the first pieces of information they are exposed to more than information presented later on \citep{asch1946forming}. The recency effect refers to the tendency for humans to more easily recall items at the end of a list compared to items in the middle of the list \citep{marshall1972effects}.

Each cell combination consists of three parts: the source 1 cell , the source 2 cell, and the target cell. We classify these cells based on their position relative to the L-shaped morphomic pattern: cells within the L-shaped morphomic pattern are labeled as \textit{In}, while those outside are labeled as \textit{Out}. For example, a combination with source 1 as digo (In), source 2 as dices (Out), and target form as diga (In) is categorized as In-Out-In.

\begin{table}[h]
\resizebox{\columnwidth}{!}{%
\begin{tabular}{@{}c|ccc|ccc|ccc@{}}
\hline
& \multicolumn{3}{c|}{\textbf{10\%L-90\%NL}} & \multicolumn{3}{c|}{\textbf{50\%L-50\%NL}} & \multicolumn{3}{c}{\textbf{90\%L-10\%NL}} \\ \hline
Cell combi. & L (\small{\%}) & NL (\small{\%}) & L/NL & L (\small{\%}) & NL (\small{\%}) & L/NL & L (\small{\%}) & NL (\small{\%}) & L/NL \\ \hline
\texttt{In-In-In} & 45.61 & 60.64 & 0.7 & 82.56 & 53.75 & 1.63 & 90.02 & 24.14 & 6.85 \\
\texttt{In-Out-Out} & 4.44 & 34.84 & 0.08 & 57.18 & 34.58 & 1.78 & 91.53 & 40.66 & 2.86 \\
\texttt{In-In-Out} & 27.17 & 62.57 & 0.4 & 63.94 & 54.19 & 1.15 & 87.66 & 22.85 & 5.07 \\
\texttt{In-Out-In} & 40.26 & 45.4 & 0.9 & 80.47 & 45.83 & 1.92 & 91.98 & 48.72 & 6.07 \\
\texttt{Out-In-In} & 40.62 & 64.77 & 0.57 & 76.17 & 57.19 & 1.41 & 89.54 & 16.93 & 15.23 \\
\texttt{Out-In-Out} & 37.68 & 66.68 & 0.54 & 70.49 & 59.58 & 1.18 & 87.43 & 16.53 & 6.58 \\
\texttt{Out-Out-In} & 30.39 & 59.23 & 0.45 & 59.8 & 52.11 & 1.08 & 87.06 & 22.42 & 7.44 \\
\texttt{Out-Out-Out} & 25.56 & 58.87 & 0.38 & 69.77 & 56.19 & 1.23 & 86.7 & 18.4 & 6.19 \\ \hline
\end{tabular}%
}
\caption{Cell combination accuracies for 10\%L-90\%NL (left), 50\%L-50\%NL (middle), and 90\%L-10\%NL (right). The mean accuracies in percentage are calculated for separately by verb types (L denotes L-shaped verbs and NL denotes NL-shaped verbs) and by cell combinations (e.g., In-In-In). L/NL denotes the ratio of the mean accuracies of the L-shaped vs NL-shaped verbs. For a visualization of this table, see Appendix \ref{sec:appendix:cellcombo}.}\label{tab:cellcombinationacc}
\end{table}

To investigate potential primacy and recency effects, we compare pairs of cell combinations where the target cell aligns with either source 1 (primacy) or source 2 (recency) (Table \ref{tab:cellcombinationacc}). For example, \textbf{In}-Out-\textbf{In} allows for a primacy effect, while Out-\textbf{In}-\textbf{In} allows for a recency effect. We evaluate these effects by identifying minimally different pairs of cell combinations, such as comparing In-Out-In (primacy) to Out-Out-In to assess the presence of a primacy effect. The results show a consistent primacy effect for L-shaped verbs in the 10\%L-90\%NL and 50\%L-50\%NL conditions, and this effect is weaker in the 90\%L-10\%NL condition. For instance, in the 10\%L-90\%NL condition, the accuracy of the In-Out-In cell combination (40.26\%) exceeds that of Out-Out-In (30.39\%), indicating an apparent primacy effect. However, we do not detect a consistent recency effect across conditions.

We also find that `In' targets are predicted more accurately than `Out' targets for L-shaped verbs across all frequency conditions, suggesting a potential bias towards these `In' cells. 

\subsection{Memorization and Generalization}\label{sec:memgen} 

We examine the models' ability to memorize and generalize morphomic patterns, focusing specifically on stem-final consonant triples under varying frequency conditions, addressing our research aim 3b. Our analysis shows that in most frequency conditions, the models show higher accuracy for memorized stem-final consonants compared to generalized ones, with the exception of the balanced 50\%L-50\%NL condition. We also find that memorization improves as the frequency of L-shaped verbs increases. 

In order to generate accurate predictions, the transformer model must balance between memorization and generalization \citep{pmlr-v70-arpit17a, zhang2021}. This balance is particularly crucial when modeling L-shaped morphomes as it involves irregular stem alternations that do not follow straightforward phonological or semantic rules. Thereby, the model must rely on memorization to reproduce the specific alternations of seen verbs. At the same time, it must be able to extend to novel alternations. We assess how varying the distribution of L-shaped verbs in the training data affects the models' ability to memorize and generalize stem-final consonant triples. We focus on triples formed by the stem-final consonants of the first source form, second source form, and target form. For example, given the forms 
(\texttt{t\textipa{R}ad\textipa{"}usen} (source 1), \texttt{t\textipa{R}adusk\textipa{"}amos} (source 2), \texttt{t\textipa{R}ad\textipa{"}uskan} (target)), the stem-final consonant triple consists of \texttt{s}, \texttt{sk}, and \texttt{sk}. We quantify memorization as the models' ability in reproducing seen stem-final consonant triples, and generalization as their ability to correctly predict unseen triples. We treat memorization and generalization as two distinct knowledge states.

Using mixed-effects logistic regression, we examine how frequency conditions and knowledge states influence prediction accuracy. Logistic mixed-effects models are implemented using the \texttt{glmer} function from the \texttt{lme4} package \citep{lme4} in \texttt{R}. Our model predicts accuracy (\texttt{prediction\_status}: correct vs. incorrect) with two fixed effects - frequency conditions and knowledge state conditions - and two random intercepts: \texttt{triples} and \texttt{model} (among 12 models). We implement the following model structure:

\begin{center}
\texttt{
glmer(prediction\_status $\sim$ knowledge\_state * frequency\_condition + (1|triples) + (1|model), \
data=df, family="binomial")
}
\end{center}

To interpret the results of our fitted model, we use the \texttt{emmeans} package \citep{emmeans} to calculate estimated marginal means (EMMs). This way, we can estimate the predicted probabilities of correct predictions for different combinations of knowledge states and frequency conditions. In the condition where the frequency distribution of verb types is equal (50\%L-50\%NL condition), we observe a slight advantage for generalization over memorization, with predicted probabilities for generalized stem-final consonant pairs being 0.113 higher than for memorized stem-final consonant pairs. However, in the 10\%L-90\%NL condition, memorized stem-final consonant pairs show a 0.036 higher predicted probability. In contrast, in the 90\%L-10\%NL condition, this advantage increases to 0.304 (see Appendix \ref{sec:appendix:memgen} for detailed results). 

In terms of memorization, we find a positive correlation between the frequency of L-shaped verbs in the training data and the probability of correct predictions. The highest probability of correct predictions occurs in the 90\%L-10\%NL condition (0.754), followed by 50\%L-50\%NL condition (0.627) and least in the 10\%L-90\%NL condition (0.252). This suggests that increased exposure to L-shaped verbs enhances the model's ability to memorize stem-final consonant triples. 

For generalization, the highest probability of correct predictions occurs in the balanced 50\%L-50\%NL condition (0.74). Interestingly, the probabilities are lower in the skewed conditions, with the 10\%L-90\%NL condition having a probability of 0.216 and the 90\%L-10\%NL condition having a probability of 0.450.

\subsection{Consonant pair analysis}\label{sec:consonant-pair-analysis} 

We investigate the models' performance on specific stem-final consonant pairs of L-shaped verbs, focusing on the alternating pairs comprising the stem-final consonant of the forms in the `Out' cells and that of the `In' cells within the paradigm (as discussed in section \ref{sec:cellcombo}), addressing our aim 3c. Our analysis shows that the models are sensitive to the input frequency of consonant alternations, indicating that they have not fully acquired the abstract morphological patterns.

For example, for the lemma \texttt{\textipa{desiR}}, the consonant pair is \texttt{\textipa{[s]-[g]}}, where \texttt{\textipa{[s]}} is the stem-final consonant of the out cells and \texttt{\textipa{[g]}} is found in forms sharing the L-shaped pattern. The most frequent pairs in the dataset are \texttt{\textipa{[s]-[sk]}}, with 141 occurrences, followed by \texttt{\textipa{[n]-[ng]}} and \texttt{\textipa{[\c{c}]-[x]}}, with 53 and 25 occurrences, respectively. This skewed distribution naturally results in varying ratios for each experimental run due to our data sampling process (as shown in Section \ref{sec:methodology}).
% This unbalanced distribution naturally leads to differing ratios for each run due to the process of data sampling and may lead to more productive generalization for frequent alternations.

We examine the models' sensitivity to consonant pair frequencies across the frequency conditions and assess how varying proportions of L-shaped in the input affect the learning of consonant alternations. Across all three runs of the 10\%L-90\%NL condition, \texttt{\textipa{[s]-[sk]}} appears frequently in both test (3-4 times) and training sets (8-13 times). Some pairs like \texttt{\textipa{[l\c{c}]-[lx]}} and \texttt{\textipa{[s]-[\textipa{g}]}} appear in test sets but are rare or absent in training, which might pose difficulty for models in applying patterns to novel combinations.
In the 50\%L-50\%NL condition, \texttt{\textipa{[s]-[sk]}} remains the most frequent pair, appearing 14-22 times in test sets and 15-23 times in training sets. Other pairs like \texttt{\textipa{[n]-[ng]}} and \texttt{\textipa{[s]-[g]}} also appear but less frequently. In the 90\%L-10\%NL condition, \texttt{\textipa{[s]-[sk]}} appears even more frequently (22-32 times in test sets and 97-104 times in training), while other pairs remain less common. Details are given in Appendix \ref{sec:appendix:consonant-pair-analysis}.
A confusion matrix in Appendix \ref{sec:appendix:confusion} summarizes the top 5 most erroneous consonant pairs  for L-shaped verbs. 

The main frequency effect can still be found consistently across these consonant pairs. In the 10\%L-90\%NL condition, \texttt{\textipa{[s]-[sk]}} achieves 68.6\% accuracy, while \texttt{\textipa{[s]-[g]}} and \texttt{\textipa{[n]-[ng]}} reach 26.2\% and 77.4\%, respectively. Accuracy increases in the 50\%L-50\%NL condition to 89.8\%, 60.6\%, and 83.6\%, respectively. In the 90\%L-10\%NL condition, accuracies further improve to 93.3\%, 92.6\%, and 91.4\%. Looking beyond accuracies, we find that the models still make systematic errors, often defaulting to more frequent lemma consonants rather than altered ones (e.g., predicting \texttt{\textipa{[s]-[s]}} instead of \texttt{\textipa{[s]-[sk]}}).

The models perform worse for less frequent or unseen alternations (such as \texttt{\textipa{[s]-[g]}}) compared to more frequent alternations (\texttt{\textipa{[s]-[sk]}}). This indicates that the models have not fully acquired the abstract morphological patterns.
In these cases, models tend to regularize L-shaped verbs in datasets, as erroneous predictions often result in non-alternating pairs. These results suggests that the models are relying heavily on frequency-based pattern matching rather than acquiring true morphological competence.

\section{Conclusion}

In this paper, we examine the learning capabilities of transformer models with respect to morphomic pattern, specifically the L-shaped pattern in Spanish. We conduct a series of post-hoc analyses to understand the factors influencing the learning of morphomic patterns under varying frequency conditions.  

Transformer models have shown remarkable ability to learn irregular patterns in related language tasks, such as English past tense inflection, German noun plurals and Arabic noun plurals \citep{kodner-khalifa-2022-sigmorphon, kakolu-ramarao-etal-2022-heimorph}. However, morphomic patterns are complex linguistic patterns that are hard to acquire \citep{Nevins2015TheRA}. The models' ability to capture morphomic patterns was not guaranteed because these patterns operate independently of semantics, syntax, and phonology. In our study, the models' performance on L-shaped verbs, especially in conditions with uneven frequency, indicates that transformers have developed some level of competence in recognizing and applying morphomic patterns from the training data.

We first look at the learning strategies of the transformer with respect to input ordering. We observe a clear primacy effect in models' processing of L-shaped verbs, suggesting that the models are more influenced by the first source cell in making predictions. We also observe that models have higher accuracy in predicting target forms that are part of L-shaped pattern compared to those outside the L-shaped pattern within a morphological paradigm. This bias suggests that the models are capturing some aspects of the overall paradigm structure, particularly the distribution of irregular forms within the L-shaped pattern. 

%This finding is similar to ordering effects in ChatGPT's processing of input lables \citep{wang2023primacy}.

In our investigation of models' strategies for balancing memorization and generalization, we show that the type frequency of L-shaped verbs impacts the models' ability to both memorize and generalize stem-final consonant alternations. For memorization, we observe a positive correlation between the frequency of L-shaped verbs in the training data and the model's ability to correctly reproduce seen stem-final consonants. This suggests that increased exposure to L-shaped verbs enhances the model's ability to retain and apply stem-final consonant alternations. This finding aligns with studies which show morphological patterns with higher type frequency to be more productive \citep[\textit{inter alia}]{bybee1995, pierrehumbert2001stochastic, baer2012role}. However, when it comes to generalization, we observe a non-linear relationship between the frequency of L-shaped verbs in the training data and the model's ability to produce unseen stem-final consonants. We find that generalization performance peaks in the balanced condition (50\%L-50\%NL), but decreases in skewed conditions (90\%L-10\%NL and 10\%L-90\%NL conditions). While increased exposure to L-shaped verbs enhances memorization, it may not necessarily lead to better generalization. 

While the models demonstrate some ability to learn and apply the L-shaped pattern, they exhibit a clear preference for regular patterns when encountering unfamiliar consonant alternations. This suggests that the models simply rely on frequency-based pattern and have not fully acquired abstract morphological rules.

Furthermore, the models' superior performance on regular verbs in the 10\%L-90\%NL condition (a close approximation to natural language distribution) validates the results from \citet{Nevins2015TheRA}'s study. In their study with Spanish speakers on a wug-test-like inflection task, 71.9\% of the participants showed a preference to NL-shaped responses over L-shaped responses. This similarity in behavior between transformers and human participants motivates a deeper comparison. In the future, we aim to compare our model's performance with the experimental results from human speakers, such as those from \citet{Nevins2015TheRA} and \citet{cappellaro2024cognitive}. Through this comparison, we could assess how effectively our computational approach captures the linguistic and cognitive phenomena observed in human morphological processing in the context of morphomic patterns. 

% can be extended to other languages \citep{Herce2023}.

\subsection*{Limitations}

We acknowledge several limitations in our current study that could be addressed in future research. First, we did not explore alternative computational approaches. For example, Recurrent Neural Networks (RNNs) have been used to model German plurals \citep{dankers2021generalising}, and Linear Discriminative Learning models have been used to model Korean verbs \citep{jeong-etal-2023-linear}, along with other approaches outlined in Section \ref{sec:related-work}.

Second, we did not perform probing analyses to investigate the internal representations of the model but instead relied solely on post-hoc analyses.

Third, to rigorously evaluate the model's generalization capabilities, it would be beneficial to use test data that is entirely unattested in the training set. This means that both the lemmas (word stems) and the feature tags (e.g., tense, number, person) should be novel to the model. This approach, similar to \citet{kodner2023morphological}, would provide a more robust assessment of the model's ability to generalize morphomic patterns to unseen data. 

Finally, we did not account for the morphological complexity of the verbs. Some verbs have prefixes, and some do not, therefore, some lemmas share the same stems. On the one hand, this renders the lemma-level train-development-test splitting procedure less effective and might artificially inflate the accuracies of our models. On the other hand, this is arguably ecologically more valid as human learners do get exposed to both morphologically complex and simple verbs.

% Enhance the ability of the model to capture abstract morphological patterns beyond frequency matching by incorporating morphological information. 

% probing for memorization - \citep{haviv-etal-2023-understanding}

\subsection*{Ethics Statement}

All the models we use are small, which significantly reduces the computational resources required for training and inference. All data used in the study are from open datasets. To promote transparency and reproducibility, all data and code used in this study are publicly available. 

The involved university does not require IRB approval for this kind of study, which uses publicly available data without involving human participants. We do not see any other concrete risks concerning dual use of our research results. Of course, in the long run, any research results on AI methods based on large language models could potentially be used in contexts of harmful and unsafe applications of AI.
But this danger is rather low in our concrete case.

\subsection*{Acknowledgements}

We gratefully acknowledge the support of the central HPC system ``HILBERT” at Heinrich Heine University Düsseldorf. We thank the attendees of the Mediterranean Morphology Meeting (2022), the Linguistics Society of America (2023), and the Linguistics Association of Great Britain conference (2024) for their valuable feedback. 
We also thank the ARR reviewers and ACs for their constructive feedback, which improved our work.
\section*{CRediT authorship contribution statement}

We follow the CRediT taxonomy\footnote{ \url{https://credit.niso.org/}}.
Conceptualization: [AKR, KT, DB]; Data curation: [AKR, KT]; Formal Analysis: [AKR]; Investigation: [AKR, KT, DB]; Methodology: [KT, DB, AKR]; Supervision: [KT, DB]; Visualization: [AKR]; and Writing – original draft: [AKR] and Writing – review \& editing: [KT, DB, AKR].

% \subsection*{Aknowledgements}
% We thank the audience of 
% the [ANON] and [ANON] for their valuable feedback.  
% We gratefully acknowledge the support of the central HPC system at [ANON].

\bibliography{custom}

\appendix

\section{Appendices}
\label{sec:appendix}

\subsection{ Accuracies}\label{sec:appendix:accuracies}

% \begin{figure}[h]
% \centering
%   \scalebox{0.7}
%     {\input{plots/overall-accuracies}}
%   \caption{Mean and confidence intervals of overall sequence accuracies across frequency conditions}\label{fig: overall-accuracy}
% \end{figure}

\begin{figure}[H]
\centering
  \scalebox{0.7}{\input{plots/stem-accuracies}}
  \caption{Mean and 95\% confidence intervals of stem sequence accuracies for L-shaped and NL-shaped verbs  across frequency conditions. {\color{gray}Gray}: NL-shaped, {\color{orange}Orange}: L-shaped.}
  \label{fig: stem-accuracies}
\end{figure}

% \begin{figure}[H]
% \centering
%   \scalebox{0.7}{\input{plots/naacl25_1_l_vs_nl_accuracy_without_stress}}
%   \caption{Mean accuracies and confidence intervals of overall sequence accuracies for L-shaped and NL-shaped verbs across frequency conditions, evaluated using the model with a batch size of 512. {\color{gray}Gray}: NL-shaped, {\color{orange}Orange}: L-shaped.}
%   \label{fig: lvsnl-512}
% \end{figure}

% \begin{figure}[H]
% \centering
%   \scalebox{0.7}{\input{plots/naacl25_l_vs_nl_accuracy_without_stress}}
%   \caption{Mean accuracies and confidence intervals of overall sequence accuracies for L-shaped and NL-shaped verbs across frequency conditions, evaluated using the model with a batch size of 800. {\color{gray}Gray}: NL-shaped, {\color{orange}Orange}: L-shaped.}
%   \label{fig: lvsnl-800}
% \end{figure}

\subsection{Cell combinations}\label{sec:appendix:cellcombo}

\begin{figure}[H]
  \centering
  \scalebox{0.7}{\input{plots/cell-combinations-10L-90NL}}
    \caption{Cell combination accuracies for 10L-90NL condition. {\color{gray}Gray}: NL-shaped, {\color{orange}Orange}: L-shaped.}
    \label{fig:cell-accuracies-10-90}
\end{figure}

\begin{figure}[H]
  \centering
  \scalebox{0.7}{\input{plots/cell-combinations-50L-50NL}}
    \caption{Cell combination accuracies for 50L-50NL condition. {\color{gray}Gray}: NL-shaped, {\color{orange}Orange}: L-shaped.}
    \label{fig:cell-accuracies-50-50}
\end{figure}

\begin{figure}[H]
  \centering
  \scalebox{0.7}{\input{plots/cell-combinations-90L-10NL}}
    \caption{Cell combination accuracies for 90L-10NL condition. {\color{gray}Gray}: NL-shaped, {\color{orange}Orange}: L-shaped.}
    \label{fig:cell-accuracies-90-10}
\end{figure}

\subsection{Memorization and Generalization task}\label{sec:appendix:memgen}

% \begin{center}
% \texttt{emmeans(model, $\sim$ phenomena\textsuperscript{*}condition, type="response")}
% \end{center}

\begin{table}[h]
\scalebox{0.6}{%
\begin{tabular}{lllllll}
\hline
Knowledge state & Condition & Prob. & SE  & asymp.LCL & asymp.UCL \\
\hline
Generalization & 10\%L-90\%NL & 0.216 & 0.0150 & 0.188 & 0.247 \\
Memorization & 10\%L-90\%NL & 0.252 & 0.0183 & 0.218 & 0.290 \\
Generalization & 50\%L-50\%NL & 0.740 & 0.0260 & 0.686 & 0.788 \\
Memorization & 50\%L-50\%NL & 0.627 & 0.0351 & 0.556 & 0.693 \\
Generalization & 90\%L-10\%NL & 0.450 & 0.0353  & 0.382 & 0.520 \\
Memorization & 90\%L-10\%NL & 0.754 & 0.0301 & 0.690 & 0.808 \\
\hline
\end{tabular}
}
\caption{Estimated marginal means for knowledge state and frequency condition combinations with asymptotic 95\% confidence intervals. `Prob.' represents the estimated marginal means, `SE' is the standard error, `asymp.LCL' is the asymptotic lower confidence limit, and `asymp.UCL' is the asymptotic upper confidence limit.}
\label{tab:emm}
\end{table}

\begin{table}[H]
\resizebox{\columnwidth}{!}{%
\begin{tabular}{lrrrrr}
\hline
\textbf{Memorization} \\
\hline
Contrast & odds.ratio & SE & z.ratio & p.value \\
\hline
10\%L-90\%NL / 50\%L-50\%NL & 0.2009 & 0.02336   & -13.806 & $<$.0001 \\
10\%L-90\%NL / 90\%L-10\%NL & 0.1100 & 0.01394   & -17.420 & $<$.0001 \\
50\%L-50\%NL / 90\%L-10\%NL & 0.5476 & 0.08795  & -3.749 & 0.0005 \\
\hline
\textbf{Generalization} \\
\hline
Contrast & odds.ratio & SE & z.ratio & p.value \\
\hline
10\%L-90\%NL / 50\%L-50\%NL & 0.0969 & 0.00949  &  -23.833 & $<$.0001 \\
10\%L-90\%NL / 90\%L-10\%NL & 0.3369 & 0.03802   & -9.642 & $<$.0001 \\
50\%L-50\%NL / 90\%L-10\%NL & 3.4750 & 0.47015   & 9.207 & $<$.0001 \\
\hline
\end{tabular}
}
\caption{Pairwise comparisons of frequency condition levels for each knowledge state. `odds.ratio' represents the estimated odds ratio, `SE' is the standard error, `z.ratio' is the z-statistic, and `p.value' is the p-value.}
\label{tab:pairs-phenomena}
\end{table}

\subsection{Consonant-pair analysis}\label{sec:appendix:consonant-pair-analysis}

\begin{table}[H]
\centering
\resizebox{0.45\textwidth}{!}{%
\begin{tabular}{ll}
\hline
Stem-final consonants & No. of occurrences \\
\hline
{[}s{]}-{[}sk{]} & 141 \\
{[}n{]}-{[}n\textipa{g}{]} & 53 \\
{[}\c{c}{]}-{[}x{]} & 25 \\
{[}s{]}-{[}\textipa{g}{]} & 15 \\
{[} {]}-{[}j\textipa{g}{]} & 14 \\
{[}\textipa{R}\c{c}{]}-{[}\textipa{R}x{]} & 10 \\
{[}n\c{c}{]}-{[}nx{]} & 10 \\
{[}l{]}-{[}l\textipa{g}{]} & 4 \\
{[}l\c{c}{]}-{[}lx{]} & 4 \\
{[}s{]}-{[}s\textipa{g}{]} & 2 \\
{[}b{]}-{[}p{]} & 1 \\
\hline
\end{tabular}
}
\caption{\label{tab: num-stem-final-consonant-pairs} Number of alternating stem final-consonant pairs of all the L-shaped verbs. `{[} {]}' represents the absence of stem-final consonant.}
\end{table}

\begin{table}[!htbp]
\centering
\scalebox{0.9}{%
\begin{tabular}{@{}cccc@{}}
\hline
& Consonant-pairs & Freq. in Test & Freq. in Train \\
\hline
\parbox[t]{5mm}{\multirow{4}{*}{\rotatebox[origin=c]{90}{\textbf{Run 1}}}}
& [s]-[sk] & 3 & 8 \\
& [n]-[n\textipa{g}] & 2 & 4 \\
& [n\c{c}]-[nx] & 1 & 1 \\
& [l\c{c}]-[lx] & 1 & 0 \\
\hline
\parbox[t]{5mm}{\multirow{4}{*}{\rotatebox[origin=c]{90}{\textbf{Run 2}}}}
& [s]-[sk] & 3 & 13 \\
& [n]-[n\textipa{g}] & 1 & 2 \\
& [n\c{c}]-[nx] & 1 & 1 \\
& [s]-[\textipa{g}] & 1 & 0 \\
\hline
\parbox[t]{5mm}{\multirow{2}{*}{\rotatebox[origin=c]{90}{\textbf{Run 3}}}}
& [s]-[sk] & 4 & 10 \\
& [s]-[\textipa{g}] & 2 & 1 \\
\hline
\end{tabular}
}
\caption{Frequency of stem-final consonants in train and test in 10\%L-90\%NL condition.}
\label{tab:freq-train-test-10L-90NL}
\end{table}

\begin{table}[H]
\centering
\scalebox{0.9}{%
\begin{tabular}{cccc}
\hline
& Consonant-pairs & Freq. in Test & Freq. in Train \\
\hline
\parbox[t]{5mm}{\multirow{7}{*}{\rotatebox[origin=c]{90}{\textbf{Run 1}}}}
& [s]-[sk] & 19 & 20 \\
& [n]-[n\textipa{g}] & 3 & 10 \\
& [n\c{c}]-[nx] & 2 & 0 \\
& [s]-[\textipa{g}] & 3 & 2 \\
& [l]-[l\textipa{g}] & 1 & 0 \\
& [ ]-[j\textipa{g}] & 3 & 0 \\
& [l\c{c}]-[lx] & 1 & 1 \\
\hline
\parbox[t]{5mm}{\multirow{7}{*}{\rotatebox[origin=c]{90}{\textbf{Run 2}}}}
& [s]-[sk] & 22 & 15 \\
& [\c{c}]-[x] & 2 & 5 \\
& [n]-[n\textipa{g}] & 3 & 7 \\
& [n\c{c}]-[nx] & 1 & 1 \\
& [l\c{c}]-[lx] & 1 & 1 \\
& [s]-[\textipa{g}] & 2 & 3 \\
& [s]-[s\textipa{g}] & 1 & 0 \\
\hline
\parbox[t]{5mm}{\multirow{8}{*}{\rotatebox[origin=c]{90}{\textbf{Run 3}}}}
& [s]-[sk] & 14 & 23 \\
& [s]-[\textipa{g}] & 4 & 0 \\
& [n]-[n\textipa{g}] & 2 & 12 \\
& [ ]-[j\textipa{g}] & 3 & 4 \\
& [l]-[l\textipa{g}] & 1 & 1 \\
& [\textipa{R}\c{c}]-[\textipa{R}x] & 1 & 2 \\
& [n\c{c}]-[nx] & 2 & 1 \\
& [\c{c}]-[x] & 2 & 4 \\
\hline
\end{tabular}
}
\caption{Frequency of stem-final consonants in train and test in 50\%L-50\%NL condition. `{[} {]}' represents the absence of stem-final consonant.}
\label{tab:freq-train-test-50L-50NL}
\end{table}

\begin{table}[H]
\centering
\scalebox{0.9}{%
\begin{tabular}{cccc}
\hline
& Consonant-pairs & Freq. in Test & Freq. in Train \\
\hline
\parbox[t]{5mm}{\multirow{8}{*}{\rotatebox[origin=c]{90}{\textbf{Run 1}}}}
& [s]-[sk] & 32 & 97 \\
& [\c{c}]-[x] & 5 & 18 \\
& [ ]-[j\textipa{g}] & 5 & 5 \\
& [n\c{c}]-[nx] & 2 & 7 \\
& [\textipa{R}\c{c}]-[\textipa{R}x] & 3 & 6 \\
& [n]-[n\textipa{g}] & 6 & 40 \\
& [l\c{c}]-[lx] & 1 & 2 \\
& [s]-[\textipa{g}] & 2 & 11 \\
\hline
\parbox[t]{5mm}{\multirow{8}{*}{\rotatebox[origin=c]{90}{\textbf{Run 2}}}}
& [s]-[sk] & 22 & 104 \\
& [\c{c}]-[x] & 2 & 5 \\
& [\textipa{R}\c{c}]-[\textipa{R}x] & 1 & 7 \\
& [s]-[\textipa{g}] & 4 & 10 \\
& [l]-[l\textipa{g}] & 2 & 2 \\
& [ ]-[j\textipa{g}] & 4 & 8 \\
& [n\c{c}]-[nx] & 3 & 7 \\
& [s]-[s\textipa{g}] & 1 & 1 \\
\hline
\parbox[t]{5mm}{\multirow{10}{*}{\rotatebox[origin=c]{90}{\textbf{Run 3}}}}
& [s]-[sk] & 25 & 103 \\
& [n\c{c}]-[nx] & 4 & 6 \\
& [s]-[\textipa{g}] & 2 & 11 \\
& [\c{c}]-[x] & 7 & 18 \\
& [l\c{c}]-[lx] & 2 & 1 \\
& [n]-[n\textipa{g}] & 10 & 35 \\
& [l]-[l\textipa{g}] & 2 & 2 \\
& [\textipa{R}\c{c}]-[\textipa{R}x] & 1 & 6 \\
& [ ]-[j\textipa{g}] & 1 & 9 \\
\hline
\end{tabular}
}
\caption{Frequency of stem-final consonants in train and test in 90\%L-10\%NL condition. `{[} {]}' represents the absence of stem-final consonant.}
\label{tab:freq-train-test-90L-10NL}
\end{table}

\begin{figure}
\begin{tikzpicture}
    \begin{axis}[
        width=\columnwidth, height=9cm,
        xlabel={\textbf{Batch sizes}},
        ylabel={\textbf{L-shaped Accuracy (\%)}},
        xtick=data,
        xticklabels={32, 64, 128, 256, 400, 512, 800, 3600},
        legend pos=north east,
        grid=major,
        ymin=30, ymax=100,
        ytick={30, 40, 50, 60, 70, 80, 90, 100},
        legend pos=south west,
        x tick label style={font=\small},
        y tick label style={font=\small},
        xlabel style={font=\bfseries},
        ylabel style={font=\bfseries},
        legend style={font=\small, draw=none, fill=none},
        title style={font=\bfseries, yshift=-2ex}
    ]
        % Plot for l_shaped_accuracy_10L_90NL
        \addplot[color=blue, mark=o, thick] coordinates {
            (1, 61.88) (2, 61.61) (3, 64.71) (4, 46.27) 
            (5, 36.75) (6, 34.86) (7, 34.17) (8, 46.74)
        };
        \addlegendentry{10\%L-90\%NL}

        % Plot for l_shaped_accuracy_50L_50NL
        \addplot[color=red, mark=square*, thick] coordinates {
            (1, 87.42) (2, 86.16) (3, 84.94) (4, 81.14) 
            (5, 72.31) (6, 75.73) (7, 82.67) (8, 80.35)
        };
        \addlegendentry{50\%L-50\%NL}

        % Plot for l_shaped_accuracy_90L_10NL
        \addplot[color=green!70!black, mark=triangle*, thick] coordinates {
            (1, 88.62) (2, 89.24) (3, 87.18) (4, 87.61) 
            (5, 88.75) (6, 87.7) (7, 88.43) (8, 86.13)
        };
        \addlegendentry{90\%L-10\%NL}
    \end{axis}
\end{tikzpicture}
    \caption{Effect of batch sizes on L-shaped accuracy for different frequency conditions (10\%L-90\%NL, 50\%L-50\%NL, and 90\%L-10\%NL).}
    \label{fig:batch-sizes}
\end{figure}

\clearpage

\subsection{Confusion Matrix}\label{sec:appendix:confusion}

\begin{table}[H]
\resizebox{\textwidth}{!}{%
\begin{tabular}{@{}lllllllllll@{}}
\hline 
\textbf{10\%L-90\%NL condition} \\
\hline
& P [s]-[s] & P [s]-[sk] & P [s]-[\textipa{g}] & P [n]-[n\textipa{g}] & P [s]-[] & P [s]-[d] & P [s]-[s\textipa{g}] & P [n]-[n] & P [n]-[\textipa{R}b] & Accuracy  \\
\hline
G [s]-[sk] & 303 & \textbf{857} & 0 & 0 & 48 & 41 & 0 & 0 & 0 & 68.61\% \\
G [s]-[\textipa{g}] & 235 & 21 & \textbf{99} & 0 & 8 & 0 & 15 & 0 & 0 & 26.19\% \\
G [n]-[n\textipa{g}] & 0 & 0 & 0 & \textbf{281} & 0 & 0 & 0 & 56 & 26 & 77.41\% \\
\hline
\textbf{50\%L-50\%NL condition}  \\
\hline
 & P [s]-[sk] & P [s]-[\textipa{g}] & P [n]-[n\textipa{g}]  & P [s]-[s] & P [s]-[d] & P [s]-[\textipa{\textesh}] & P [s]-[\textipa{j\textipa{g}}] & P [n]-[n] & P [n]-[mp] & Accuracy  \\
\hline
G [s]-[sk] & \textbf{6170} & 32 & 0 & 523 & 73 & 70 & 0 & 0 & 0 & 89.84\% \\
G [s]-[\textipa{g}] & 112 & \textbf{613} & 0 & 210 & 0 & 0 & 76 & 0 & 0 & 60.63\% \\
G [n]-[n\textipa{g}] & 0 & 0 & \textbf{853} & 0 & 0 & 0 & 0 & 139 & 29 & 83.55\% \\
\hline
\textbf{90\%L-10\%NL condition} \\
\hline
 & P [s]-[sk] & P [n]-[n\textipa{g}] & P [\c{c}]-[x] & P [s]-[s] & P [s]-[\textipa{R}s] & P [n]-[n] & P [n]-[sk] & P [n]-[n] & P [\c{c}]-[\c{c}] & Accuracy   \\
\hline
G [s]-[sk] & \textbf{9473} & 0 & 0 & 570 & 36 & 70 & 0 & 0 & 0 & 93.34\% \\
G [n]-[n\textipa{g}] & 0 & \textbf{3394} & 0 & 0 & 0 & 130 & 0 & 139 & 0 & 92.66\% \\
G [\c{c}]-[x] & 0 & 0 & \textbf{1954} & 0 & 0 & 0 & 0 & 0 & 183 & 91.44\% \\
\hline
\end{tabular}%
}
\caption{\label{tab:90L-10NL-cp-analysis} Confusion matrix for the top 3 most erroneous consonant-pairs (considering mean values) for L-shaped verbs across conditions. G = Gold, P = Prediction. Bold indicates the correct predictions.}
\end{table}

\end{document}

%% file: plots/lvsnl-accuracy-without-stress.tex
\begin{tikzpicture}

\definecolor{darkgray176}{RGB}{176,176,176}
\definecolor{gray}{RGB}{128,128,128}
\definecolor{orange}{RGB}{255,165,0}
\definecolor{steelblue33135187}{RGB}{33,135,187}
\definecolor{darkblue}{RGB}{0,0,139}
\definecolor{green}{RGB}{0,128,0}
\definecolor{lightorange}{RGB}{255,200,150}

\begin{axis}[
tick align=outside,
tick pos=left,
x grid style={darkgray176},
xmin=0.7625, xmax=3.2375,
xtick style={color=black},
xtick={1,2,3},
xticklabels={\small{10\%L-90\%NL},\small{50\%L-50\%NL},\small{90\%L-10\%NL}},
x tick label style={align=center, text width=2cm},
y grid style={darkgray176},
ylabel={\textbf{Accuracy (in \%)}},
xlabel={\textbf{Conditions}},
ymin=0, ymax=94.2331747994505,
ytick style={color=black},
legend style={at={(0.05,0.05)}, anchor=south west, font=\small},
legend cell align={left}
]

% NL-shaped verbs (gray dot)
\addplot[
    only marks,
    mark=*,
    mark size=4pt,
    color=gray,
] coordinates {
    (1.00,68.3333)
};

% L-shaped verbs (orange dot)
\addplot[
    only marks,
    mark=*,
    mark size=4pt,
    color=orange,
] coordinates {
    (1.00,56.5833)
};

\legend{NL-shaped, L-shaped}

% First set of confidence intervals (blue, solid)
\addplot [ultra thick, steelblue33135187]
table {%
1 46.3918520916791
1 66.7748145749876
};
\addplot [ultra thick, steelblue33135187]
table {%
0.875 46.3918520916791
1.125 46.3918520916791
};
\addplot [ultra thick, steelblue33135187]
table {%
0.875 66.7748145749876
1.125 66.7748145749876
};

% Second set of confidence intervals (dark blue, dashed)
\addplot [ultra thick, darkblue, dashed]
table {%
1 64.2437683914182
1 72.4228982752485
};
\addplot [ultra thick, darkblue, dashed]
table {%
0.875 64.2437683914182
1.125 64.2437683914182
};
\addplot [ultra thick, darkblue, dashed]
table {%
0.875 72.4228982752485
1.125 72.4228982752485
};

% Third set (blue, solid)
\addplot [ultra thick, steelblue33135187]
table {%
2 68.5593085271784
2 85.7740248061549
};
\addplot [ultra thick, steelblue33135187]
table {%
1.875 68.5593085271784
2.125 68.5593085271784
};
\addplot [ultra thick, steelblue33135187]
table {%
1.875 85.7740248061549
2.125 85.7740248061549
};

% Fourth set (dark blue, dashed)
\addplot [ultra thick, darkblue, dashed]
table {%
2 66.053063517671
2 73.7802698156624
};
\addplot [ultra thick, darkblue, dashed]
table {%
1.875 66.053063517671
2.125 66.053063517671
};
\addplot [ultra thick, darkblue, dashed]
table {%
1.875 73.7802698156624
2.125 73.7802698156624
};

% Fifth set (blue, solid)
\addplot [ultra thick, steelblue33135187]
table {%
3 87.9646350744707
3 91.0353649255293
};
\addplot [ultra thick, steelblue33135187]
table {%
2.875 87.9646350744707
3.125 87.9646350744707
};
\addplot [ultra thick, steelblue33135187]
table {%
2.875 91.0353649255293
3.125 91.0353649255293
};

% Sixth set (dark blue, dashed)
\addplot [ultra thick, darkblue, dashed]
table {%
3 27.0791674471038
3 35.0874992195629
};
\addplot [ultra thick, darkblue, dashed]
table {%
2.875 27.0791674471038
3.125 27.0791674471038
};
\addplot [ultra thick, darkblue, dashed]
table {%
2.875 35.0874992195629
3.125 35.0874992195629
};

% Orange dots
\addplot [ultra thick, orange, mark=*, mark size=3, mark options={solid}, only marks]
table {%
1 56.5833333333333
2 77.1666666666667
3 89.5
};

% Gray dots
\addplot [ultra thick, gray, mark=*, mark size=3, mark options={solid}, only marks]
table {%
1 68.3333333333333
2 69.9166666666667
3 31.0833333333333
};

\end{axis}

\end{tikzpicture}

%% file: plots/stem-accuracies.tex
% This file was created with tikzplotlib v0.10.1.
\begin{tikzpicture}

\definecolor{darkgray176}{RGB}{176,176,176}
\definecolor{gray}{RGB}{128,128,128}
\definecolor{orange}{RGB}{255,165,0}
\definecolor{steelblue33135187}{RGB}{33,135,187}

\begin{axis}[
tick align=outside,
tick pos=left,
x grid style={darkgray176},
xmin=0.7625, xmax=3.2375,
xtick style={color=black},
xtick={1,2,3},
xticklabels={\small{10\%L-90\%NL},\small{50\%L-50\%NL},\small{90\%L-10\%NL}},
x tick label style={align=center, text width=2cm},
y grid style={darkgray176},
ylabel={\textbf{Accuracy (in \%)}},
xlabel={\textbf{Conditions}},
ymin=0, ymax=94.2331747994505,
ytick style={color=black},
legend style={at={(0.05,0.05)}, anchor=south west, font=\small},
legend cell align={left}
]

\addplot [ultra thick, steelblue33135187]
table {%
1 46.3918520916791
1 66.7748145749876
};
\addplot [ultra thick, steelblue33135187]
table {%
0.875 46.3918520916791
1.125 46.3918520916791
};
\addplot [ultra thick, steelblue33135187]
table {%
0.875 66.7748145749876
1.125 66.7748145749876
};
\addplot [ultra thick, orange, mark=*, mark size=3, mark options={solid}, only marks]
table {%
1 56.5833333333333
};
\addplot [ultra thick, steelblue33135187]
table {%
1 64.2437683914182
1 72.4228982752485
};
\addplot [ultra thick, steelblue33135187]
table {%
0.875 64.2437683914182
1.125 64.2437683914182
};
\addplot [ultra thick, steelblue33135187]
table {%
0.875 72.4228982752485
1.125 72.4228982752485
};
\addplot [ultra thick, gray, mark=*, mark size=3, mark options={solid}, only marks]
table {%
1 68.3333333333333
};
\addplot [ultra thick, steelblue33135187]
table {%
2 68.5593085271784
2 85.7740248061549
};
\addplot [ultra thick, steelblue33135187]
table {%
1.875 68.5593085271784
2.125 68.5593085271784
};
\addplot [ultra thick, steelblue33135187]
table {%
1.875 85.7740248061549
2.125 85.7740248061549
};
\addplot [ultra thick, orange, mark=*, mark size=3, mark options={solid}, only marks]
table {%
2 77.1666666666667
};
\addplot [ultra thick, steelblue33135187]
table {%
2 66.053063517671
2 73.7802698156624
};
\addplot [ultra thick, steelblue33135187]
table {%
1.875 66.053063517671
2.125 66.053063517671
};
\addplot [ultra thick, steelblue33135187]
table {%
1.875 73.7802698156624
2.125 73.7802698156624
};
\addplot [ultra thick, gray, mark=*, mark size=3, mark options={solid}, only marks]
table {%
2 69.9166666666667
};
\addplot [ultra thick, steelblue33135187]
table {%
3 87.9646350744707
3 91.0353649255293
};
\addplot [ultra thick, steelblue33135187]
table {%
2.875 87.9646350744707
3.125 87.9646350744707
};
\addplot [ultra thick, steelblue33135187]
table {%
2.875 91.0353649255293
3.125 91.0353649255293
};
\addplot [ultra thick, orange, mark=*, mark size=3, mark options={solid}, only marks]
table {%
3 89.5
};
\addplot [ultra thick, steelblue33135187]
table {%
3 27.0791674471038
3 35.0874992195629
};
\addplot [ultra thick, steelblue33135187]
table {%
2.875 27.0791674471038
3.125 27.0791674471038
};
\addplot [ultra thick, steelblue33135187]
table {%
2.875 35.0874992195629
3.125 35.0874992195629
};
\addplot [ultra thick, gray, mark=*, mark size=3, mark options={solid}, only marks]
table {%
3 31.0833333333333
};
\end{axis}

\end{tikzpicture}

%% file: plots/cell-combinations-10L-90NL.tex
% This file was created with tikzplotlib v0.10.1.
\begin{tikzpicture}

\definecolor{darkgray176}{RGB}{176,176,176}
\definecolor{gray}{RGB}{128,128,128}
\definecolor{lightgray204}{RGB}{204,204,204}
\definecolor{orange}{RGB}{255,165,0}

\begin{axis}[
legend cell align={left},
legend style={fill opacity=0.8, draw opacity=1, text opacity=1, draw=lightgray204},
tick align=outside,
tick pos=left,
x grid style={darkgray176},
xlabel={},
xmin=-0.75, xmax=7.5,
xtick style={color=black},
xtick={0,1,2,3,4,5,6,7},
xticklabel style={rotate=90.0},
xticklabels={
  In-In-In,
  In-Out-Out,
  In-In-Out,
  In-Out-In,
  Out-In-In,
  Out-In-Out,
  Out-Out-In,
  Out-Out-Out
},
y grid style={darkgray176},
ylabel={Accuracy (in \%)},
ymin=0, ymax=70.014,
ytick style={color=black}
]
\draw[draw=none,fill=orange] (axis cs:-0.375,0) rectangle (axis cs:-0.125,45.61);
\draw[draw=none,fill=orange] (axis cs:0.625,0) rectangle (axis cs:0.875,4.44);
\draw[draw=none,fill=orange] (axis cs:1.625,0) rectangle (axis cs:1.875,27.17);
\draw[draw=none,fill=orange] (axis cs:2.625,0) rectangle (axis cs:2.875,40.26);
\draw[draw=none,fill=orange] (axis cs:3.625,0) rectangle (axis cs:3.875,40.62);
\draw[draw=none,fill=orange] (axis cs:4.625,0) rectangle (axis cs:4.875,37.68);
\draw[draw=none,fill=orange] (axis cs:5.625,0) rectangle (axis cs:5.875,30.39);
\draw[draw=none,fill=orange] (axis cs:6.625,0) rectangle (axis cs:6.875,25.56);
\draw[draw=none,fill=gray] (axis cs:-0.125,0) rectangle (axis cs:0.125,60.64);
\draw[draw=none,fill=gray] (axis cs:0.875,0) rectangle (axis cs:1.125,34.84);
\draw[draw=none,fill=gray] (axis cs:1.875,0) rectangle (axis cs:2.125,62.57);
\draw[draw=none,fill=gray] (axis cs:2.875,0) rectangle (axis cs:3.125,45.4);
\draw[draw=none,fill=gray] (axis cs:3.875,0) rectangle (axis cs:4.125,64.77);
\draw[draw=none,fill=gray] (axis cs:4.875,0) rectangle (axis cs:5.125,66.68);
\draw[draw=none,fill=gray] (axis cs:5.875,0) rectangle (axis cs:6.125,59.23);
\draw[draw=none,fill=gray] (axis cs:6.875,0) rectangle (axis cs:7.125,58.87);
\end{axis}

\end{tikzpicture}

%% file: plots/cell-combinations-50L-50NL.tex
% This file was created with tikzplotlib v0.10.1.
\begin{tikzpicture}

\definecolor{darkgray176}{RGB}{176,176,176}
\definecolor{gray}{RGB}{128,128,128}
\definecolor{lightgray204}{RGB}{204,204,204}
\definecolor{orange}{RGB}{255,165,0}

\begin{axis}[
legend cell align={left},
legend style={fill opacity=0.8, draw opacity=1, text opacity=1, draw=lightgray204},
tick align=outside,
tick pos=left,
x grid style={darkgray176},
xlabel={},
xmin=-0.75, xmax=7.5,
xtick style={color=black},
xtick={0,1,2,3,4,5,6,7},
xticklabel style={rotate=90.0},
xticklabels={
  In-In-In,
  In-Out-Out,
  In-In-Out,
  In-Out-In,
  Out-In-In,
  Out-In-Out,
  Out-Out-In,
  Out-Out-Out
},
y grid style={darkgray176},
ylabel={Accuracy (in \%)},
ymin=0, ymax=86.688,
ytick style={color=black}
]
\draw[draw=none,fill=orange] (axis cs:-0.375,0) rectangle (axis cs:-0.125,82.56);
\draw[draw=none,fill=orange] (axis cs:0.625,0) rectangle (axis cs:0.875,57.18);
\draw[draw=none,fill=orange] (axis cs:1.625,0) rectangle (axis cs:1.875,63.94);
\draw[draw=none,fill=orange] (axis cs:2.625,0) rectangle (axis cs:2.875,80.47);
\draw[draw=none,fill=orange] (axis cs:3.625,0) rectangle (axis cs:3.875,76.17);
\draw[draw=none,fill=orange] (axis cs:4.625,0) rectangle (axis cs:4.875,70.49);
\draw[draw=none,fill=orange] (axis cs:5.625,0) rectangle (axis cs:5.875,59.8);
\draw[draw=none,fill=orange] (axis cs:6.625,0) rectangle (axis cs:6.875,69.77);
\draw[draw=none,fill=gray] (axis cs:-0.125,0) rectangle (axis cs:0.125,53.75);
\draw[draw=none,fill=gray] (axis cs:0.875,0) rectangle (axis cs:1.125,34.58);
\draw[draw=none,fill=gray] (axis cs:1.875,0) rectangle (axis cs:2.125,54.19);
\draw[draw=none,fill=gray] (axis cs:2.875,0) rectangle (axis cs:3.125,45.83);
\draw[draw=none,fill=gray] (axis cs:3.875,0) rectangle (axis cs:4.125,57.19);
\draw[draw=none,fill=gray] (axis cs:4.875,0) rectangle (axis cs:5.125,59.58);
\draw[draw=none,fill=gray] (axis cs:5.875,0) rectangle (axis cs:6.125,52.11);
\draw[draw=none,fill=gray] (axis cs:6.875,0) rectangle (axis cs:7.125,56.19);
\end{axis}

\end{tikzpicture}

%% file: plots/cell-combinations-90L-10NL.tex
% This file was created with tikzplotlib v0.10.1.
\begin{tikzpicture}

\definecolor{darkgray176}{RGB}{176,176,176}
\definecolor{gray}{RGB}{128,128,128}
\definecolor{lightgray204}{RGB}{204,204,204}
\definecolor{orange}{RGB}{255,165,0}

\begin{axis}[
legend cell align={left},
legend style={fill opacity=0.8, draw opacity=1, text opacity=1, draw=lightgray204},
tick align=outside,
tick pos=left,
x grid style={darkgray176},
xlabel={},
xmin=-0.75, xmax=7.5,
xtick style={color=black},
xtick={0,1,2,3,4,5,6,7},
xticklabel style={rotate=90.0},
xticklabels={
  In-In-In,
  In-Out-Out,
  In-In-Out,
  In-Out-In,
  Out-In-In,
  Out-In-Out,
  Out-Out-In,
  Out-Out-Out
},
y grid style={darkgray176},
ylabel={Accuracy (in \%)},
ymin=0, ymax=96.579,
ytick style={color=black}
]
\draw[draw=none,fill=orange] (axis cs:-0.375,0) rectangle (axis cs:-0.125,90.02);
\draw[draw=none,fill=orange] (axis cs:0.625,0) rectangle (axis cs:0.875,91.53);
\draw[draw=none,fill=orange] (axis cs:1.625,0) rectangle (axis cs:1.875,87.66);
\draw[draw=none,fill=orange] (axis cs:2.625,0) rectangle (axis cs:2.875,91.98);
\draw[draw=none,fill=orange] (axis cs:3.625,0) rectangle (axis cs:3.875,89.54);
\draw[draw=none,fill=orange] (axis cs:4.625,0) rectangle (axis cs:4.875,87.43);
\draw[draw=none,fill=orange] (axis cs:5.625,0) rectangle (axis cs:5.875,87.06);
\draw[draw=none,fill=orange] (axis cs:6.625,0) rectangle (axis cs:6.875,86.7);
\draw[draw=none,fill=gray] (axis cs:-0.125,0) rectangle (axis cs:0.125,24.14);
\draw[draw=none,fill=gray] (axis cs:0.875,0) rectangle (axis cs:1.125,40.66);
\draw[draw=none,fill=gray] (axis cs:1.875,0) rectangle (axis cs:2.125,22.85);
\draw[draw=none,fill=gray] (axis cs:2.875,0) rectangle (axis cs:3.125,48.72);
\draw[draw=none,fill=gray] (axis cs:3.875,0) rectangle (axis cs:4.125,16.93);
\draw[draw=none,fill=gray] (axis cs:4.875,0) rectangle (axis cs:5.125,16.53);
\draw[draw=none,fill=gray] (axis cs:5.875,0) rectangle (axis cs:6.125,22.42);
\draw[draw=none,fill=gray] (axis cs:6.875,0) rectangle (axis cs:7.125,18.4);
\end{axis}

\end{tikzpicture}